\documentclass[conference]{IEEEtran}
\usepackage{amssymb}
\usepackage{amsmath}
\usepackage{graphicx} 
\usepackage{times}
\usepackage{epsfig}
\usepackage{graphicx}
\usepackage{amsmath}
\usepackage{amssymb}
\usepackage{float}
\usepackage{comment}
\usepackage{lipsum}
\usepackage{caption}    
\usepackage{subfig}    
\usepackage[most]{tcolorbox}
\usepackage{hyperref}
\IEEEoverridecommandlockouts  

\title{\LARGE \bf
Design and optimal springs stiffness estimation of a Modular OmniCrawler in-pipe climbing Robot}

\author{Akash Singh$^{*1}$, Enna Sachdeva$^{*1}$, Abhishek Sarkar$^{1}$, K.Madhava Krishna$^{1}$
\thanks{*Equal Contribution}
\thanks{$^{1}$ Robotics Research Center, IIIT-Hyderabad, India}%
\thanks{{\tt\small akashvnit2016@gmail.com}}%
\thanks{{\tt\small sachdeva.enna@research.iiit.ac.in}}%
\thanks{{\tt\small abhishek.sarkar@iiit.ac.in}}%
\thanks{{\tt\small mkrishna@iiit.ac.in}}%
}

\begin{document}

\maketitle
\thispagestyle{empty}
\pagestyle{empty}

\begin{abstract}
This paper discusses the design of a novel compliant in-pipe climbing modular robot for small diameter pipes. The robot consists of a kinematic chain of 3 OmniCrawler modules with a link connected in between 2 adjacent modules via compliant joints. While the tank-like crawler mechanism provides good traction on low friction surfaces, its circular cross-section makes it holonomic. The holonomic motion assists it to re-align in a direction to avoid obstacles during motion as well as  overcome turns with a minimal energy posture. Additionally, the modularity enables it to negotiate T-junction without motion singularity. The  compliance is realized using 4 torsion springs incorporated in joints joining 3 modules with 2 links. For a desirable pipe diameter (\text{\O} 75mm), the springs' stiffness values are obtained by formulating a constraint optimization problem which has been simulated in ADAMS MSC and further validated on a real robot prototype. In order to negotiate smooth vertical bends and friction coefficient variations in pipes, the design was later modified by replacing springs with series elastic actuators (SEA) at 2 of the 4 joints.  

\end{abstract}

\section{INTRODUCTION}
Pipeline networks support our everyday way of life through the transportation of water, crude oil and other products consumed on our daily basis. Along with the necessity to increase the efficiency of transportation, ensuring its reliable and safe operation through non-destructive testing (NDT), inspection and maintenance are equally important. To facilitate these tasks in pipelines inaccessible by humans, such as oil and gas pipes buried beneath under the sea, power plants, boilers, etc., various researchers and practitioners have designed in-pipe climbing robots for small diameter pipes, which in turn reduce the inspection time as well as cost.

The locomotion mechanism of in-pipe climbing robot is broadly categorized as wheeled, wall-press, legged, inchworm and Screw robots\cite{c1}. Numerous multi-linked wheeled in-pipe climbing robots have been developed to primarily enhance the traveling performance and speed of locomotion in small diameter pipes. To overcome smooth bends, Dertien.et.al \cite{c2} proposed an omnidirectional wheeled robot for in-pipe inspection (PIRATE). Similarly,  a series of multi-link articulated snake-like robots `PipeTron' with active wheels has been proposed by Hirose \cite{c3} where PipeTron-I robot clamps in the pipe with its zig-zag curvature attained as a result of the difference in the tensions of 2 threads going through its backbone. This differential tension also creates a twisting motion to bend the robot in pipe turns. PipeTron-VII achieves turning motion by the differential speed of each driving wheel. Furthermore, a series of Multifunctional Robot for IN-pipe inSPECTion (MRINSPECT) robot  \cite{c4}, \cite{c5},  \cite{c6},  \cite{c9} has been developed for a range of pipe diameters, where MRINSPECT IV  \cite{c6} was specifically designed for pipes of \text{\O}100mm. However, the robot can not realize backward motion in the T-junction when its rear module loses contact with the pipe surface.


\begin{figure}[t]
\centering
\hspace{4cm}
\includegraphics[width=0.4\textwidth,height=0.2\textheight]{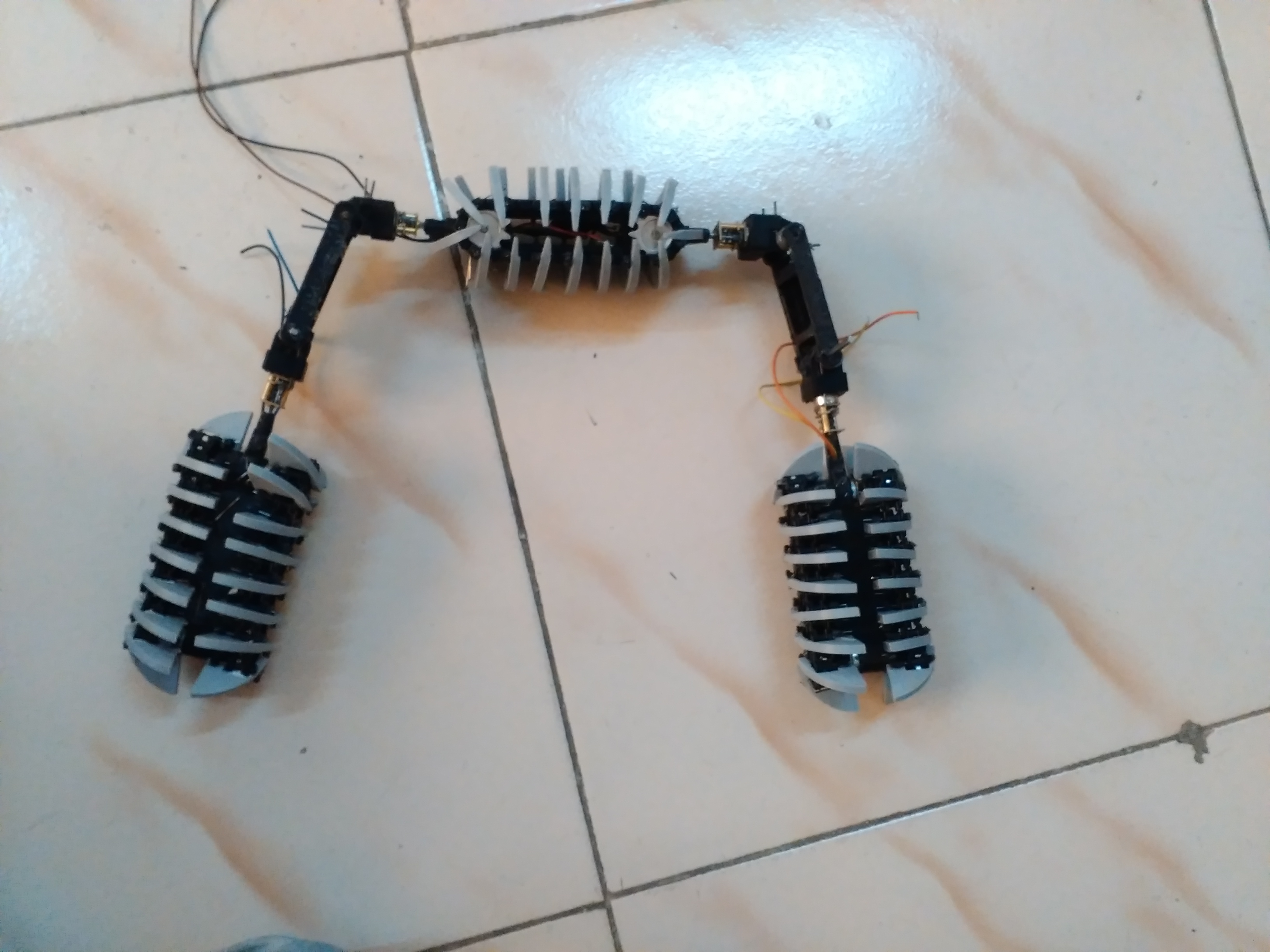}
\caption{Prototype of the Proposed Design }
\label{fig:robot_view} 
\end{figure}

Another category belongs to robots based on screw mechanism as they employ a simple transmission mechanism using a single actuator for both driving and rolling motion and provide added advantages with their efficient helical motion inside pipes. These advantages have been further augmented by Hirose \cite{c7} in `Thes-II' robot by interconnecting multiple such modules for long-distance locomotion in \text{\O}50mm gas pipelines. Atsushi \cite{c8} also designed a screw driving robot using 2 actuators with one being used for driving and rolling motion and the other to select pathways (for steering mode) in branched pipes.
 
The wheeled robots discussed so far indeed possess good maneuverability and steering abilities to overcome pipe bends but fail to drive over obstacles and often get stuck on uneven pipe surface \cite{c1}. Also, they do not provide sufficient traction force for climbing on smooth and sticky surfaces and experience the problem of wheel slip on low friction surface. Though these problems have been overcome by designing a number of tank-like crawler robots \cite{c10,c12} which provides greater traction, they suffer from motion singularity at T-junctions. While Kwon \cite{c10} tried to address this problem by connecting a series of such crawler modules where collaborative control of modules overcomes motion singularity, it leads to an increase in size and weight of the robot. Also, the conventional crawler mechanism employed in these robots rely heavily on the evenness of the inner surface of pipes and may easily get stuck in accumulated scale or obstacles in pipes\cite{c14}. Apart from that, the moving capability of previous crawler robots while making a transition from bigger to smaller diameter pipes has not been documented. This paper intends to address these limitations with the design of a modular OmniCrawler based Compliant in-Pipe climbing robot. The design of the crawler module has been inspired by a novel OmniCrawler robot \cite{c11}. 
It employs a series of circular cross-section lugs coated with rubber, which results in a significant increase in the contact area with the pipe surface as well as the traction between the lugs-pipe surface interface. The circular cross-section avoids the problem of sinking of modules in marshy surfaces. The holonomic motion of modules assists in aligning the robot along the direction of bends, beforehand. The modularity of the robot eases its maneuverabiliy with respect to variation in pipe diameter (lower to higher as well as higher to lower diameters) while exploiting the advantages of OmniCrawler modules.

This paper has been organized as follows. Section \ref{Model Description} gives a description of the robot mechanism. Section \ref{optimization} discusses the formulation to estimate optimal springs' stiffness values of the compliant joints. Furthermore, simulations and experimental results to validate the mathematical model of the design are given in Section \ref{results}.

\section{Mechanism Design}
\label{Model Description}

\begin{figure}[h!]
\centering
\hspace{-1cm}
\includegraphics[width=0.4\textwidth,height=0.10\textheight]{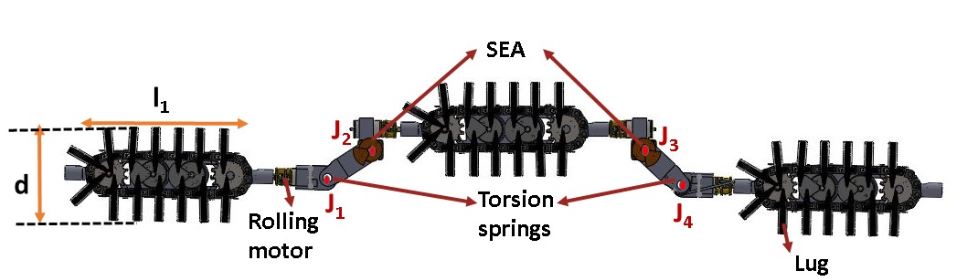}
\caption{Side view of the  CAD model of the robot}
\label{fig:robot_cad_view} 
\end{figure}

\subsection{Modular Design of the Robot}

The proposed robot has a kinematic chain of 3 OmniCrawler modules interconnected by links via compliant joints ($J_1$, $J_2$, $J_3$, $J_4$) as shown in Fig. \ref{fig:robot_cad_view}. The size of modules is determined by design constraints posed by the actuators' size and as well as  pipe environment, as explained in the subsequent sections.  

\begin{table}[h!] 
\caption{Design Parameters of the robot}
\label{table:Design Parameters}
\centering
\begin{tabular}{|l|c|r|}
\hline
\textbf{Quantity} & \textbf{Symbol} & \textbf{Values}  \\
\hline
Mass of module & $m_m$ &  0.150kg  \\
\hline
Mass of link & $m_l$ &  0.020kg  \\
\hline
Length of modules & $l_1$,$l_2$,$l_3$ & 0.14m\\
\hline
Diameter of modules & $d$ & 0.050m  \\
\hline
Length of links & $L_1$,$L_2$ & 0.060m  \\

\hline
Range of pipe diameter & $D$  & 0.065m to 0.1m  \\
\hline
Pololu Micro Motor &   &   \\
(Driving motors) saturation torque & $\tau_{max}$  & 1Nm  \\
\hline
\end{tabular}
\end{table}

\begin{figure}[h!]
\centering
\hspace{4cm}
\includegraphics[width=0.4\textwidth,height=0.2\textheight]{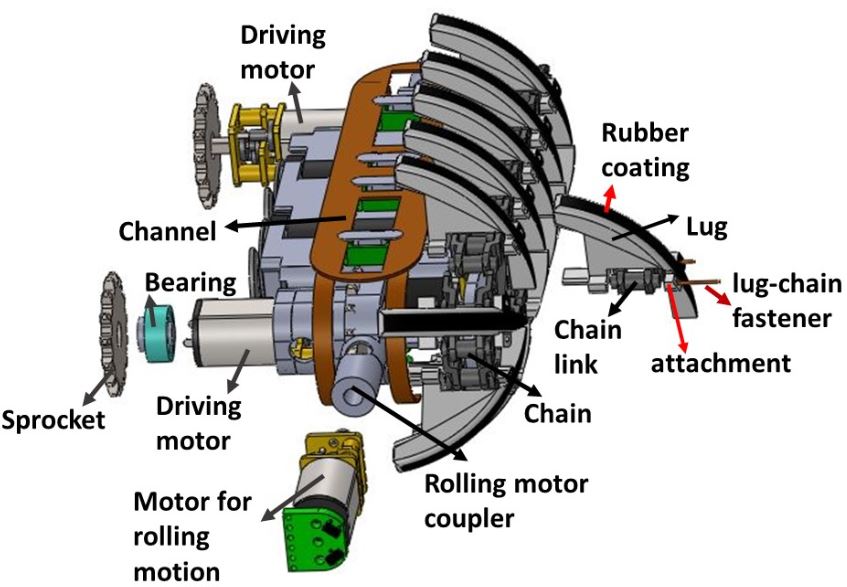}
\caption{Exploded view of the module}
\label{fig:robot_view} 
\end{figure}

\begin{figure}[htp]
  \centering
   
   \begin{minipage}[c]{0.2\textwidth}
  \hspace{0cm}  \includegraphics[width=0.9\textwidth,height=0.15\textheight]{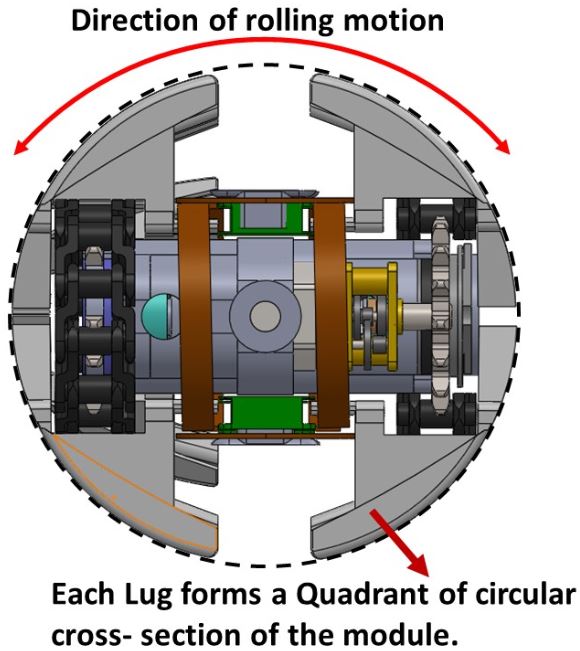}
     \caption{Cross-Sectional view of the module}
     \label{fig:cross_section}
  \end{minipage}\qquad
  \begin{minipage}[c]{0.2\textwidth}
  \includegraphics[width=1.4\textwidth,height=0.16\textheight]{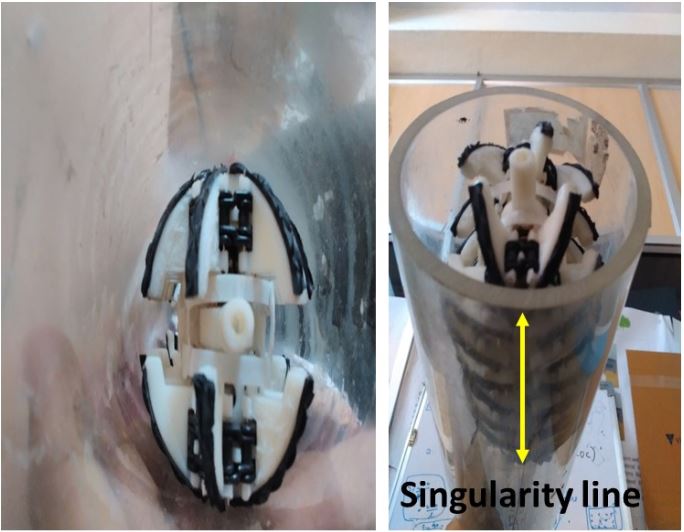}
       \caption{Module in singular configuration inside pipe}
        \label{fig:singular}
  \end{minipage}

\end{figure}

\subsubsection{OmniCrawler Module}
The OmniCrawler module incorporates a couple of chain-sprocket power transmission pairs actuated synchronously by 2 Pololu metal geared motors to provide driving motion in forward and backward direction. It is shown in the exploded view of the module assembly in Fig. \ref{fig:robot_view}. The holonomic sideways rolling motion of the robot is characterized by the circular cross-section of the module, which is achieved by the design and arrangement of the lugs. Series of lugs rest on 2 identical parallel chain links through attachments via fastener and are coated with a layer of latex rubber to provide sufficient traction for climbing, as shown in Fig. \ref{fig:robot_view} and \ref{fig:cross_section}. The sideways rolling motion is realized with an external actuator connected to the module's chassis with a coupler.

During sideways rolling, the OmniCrawler module has the ability to crawl smoothly in any configuration, except at the singularity line that does not allow any longitudinal traction force to be generated \cite{c15}. This drawback is overcome with a series of 3 such modules in the proposed design, kept at offset configuration with respect to each other. A series of modules ensures that the collaborative push/pull force generated by the modules enables robot to traverse/propagate even when any of the 3 modules touches the pipe surface at the singular line. This is illustrated in Fig. \ref{fig:singular}, where the module shown is in singular position but the robot is able to traverse forward.  
The video demonstrating this experiment can be found here : \url{https://youtu.be/D2v92VkVWl4}.

\subsubsection{Link design}
A link assembly connects 2 OmniCrawler modules via compliant joints. The design parameters of each link is determined by the pipe diameter as well as curvature at the pipe bends. The link has been designed to incorporate rolling motor clamps, torsion springs as well as SEAs. The link design parameters are specified in Table \ref{table:Design Parameters}.

\subsubsection{Series Elastic Actuator (SEA)}
\begin{figure}[h!]
\hspace{0cm}
\subfloat[Dual shaft-spring assembly]{\includegraphics[width=0.22\textwidth,height=0.12\textheight]{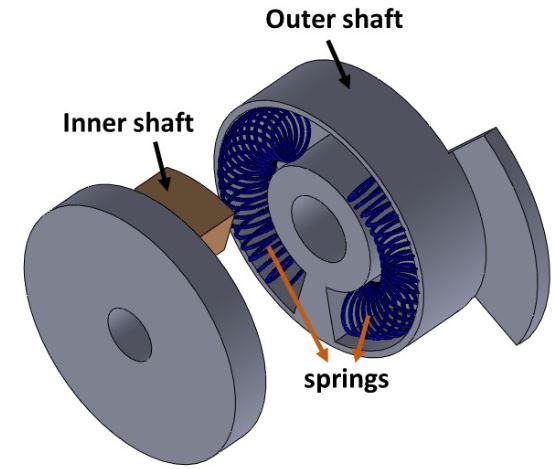}\label{fig:SEA_CAD}}
\hspace{0cm}
\subfloat[Series Elastic Actuator]{\includegraphics[width=0.22\textwidth,height=0.15\textheight]{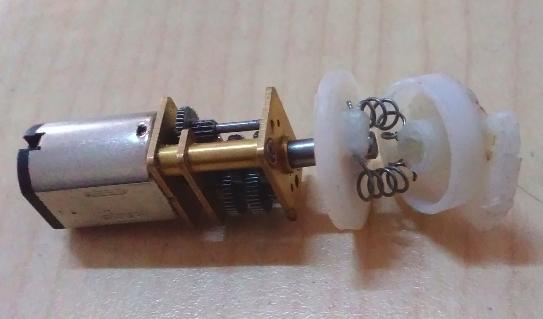}\label{fig:SEA_motor}}
\caption{Series Elastic Actuator}
\label{fig:Series Elastic Actuator}
\end{figure}

For a desirable pipe diameter (\text{\O}75mm) and friction coefficient, torsion springs designed with estimated optimal stiffness are incorporated at 4 joints ($J_1$,$J_2$,$J_3$,$J_4$). To further extend the capability of the robot to overcome smooth turns and comply with friction coefficient variations, springs at joints $J_2$ and $J_3$ are replaced by an arrangement of geared motor in series with a linear spring, called SEA \cite{c13}. Here, SEA has been designed with an assembly of dual circular shaft with linear extension springs, where the pair of springs are embedded in between these 2 shafts. The arrangement is such that the outer shaft is connected to the joints of the link and the inner shaft is connected to the motor shaft as shown in Fig. \ref{fig:Series Elastic Actuator}. The springs are encased inside circular slots of outer shaft with one of their ends attached to the outer shaft and their other ends attached to the inner shaft, as shown in Fig. \ref{fig:SEA_motor}.
When inner shaft is actuated, the ends of the springs connected to it are displaced inside the slots and the torque is transfered to the outer shaft via springs, thereby actuating the link. Encoders could later be integrated with the SEA for sensing spring deflection and estimate the joint torque for closed loop torque control. Therefore, the preloaded torsion springs at joints $J_1$, $J_4$ and SEAs at $J_2$, $J_3$ provide the necessary clamping force in a pipe and filters out vibrations while overcoming jagged terrains. The excess torque input to the SEAs are stored in the springs while complying with diameter variations, thereby protecting geared motors from getting damaged.

\begin{table}[t] 
\hspace{-2cm}
\centering
\caption{Nomenclature for model description}
\begin{tabular}{|p{2cm}|p{5cm}|}
\hline
\textbf{Symbols} & \textbf{Quantity}  \\
\hline
$k_1,k_2,k_3,k_4$ & torsion spring constant of 4 passive joints \\

\hline
$i$ & represents $i_{th}$ module , \\

\hline
$wm_{i}$ & weight of $i_{th}$ module \\

\hline
$d$  & Diameter of the module \\

\hline
$l_{i}$  &  length of module \\

\hline
$F_{i}$  &  Friction force of $i^{th}$ module\\

\hline
$N_{i}$ & Normal force acting on $i^{th}$ module\\

\hline
$\theta_{i}$  &  angle of $i^{th}$ module with global x(horizontal) axis\\

\hline
$J_{i}$  &  represents $i^{t}$ joint  \\

\hline
$wl_k$ & weight of $k^{th}$ link ( $k^{th}$ link connects  $k^{th}$ module with  $k+1^{th}$ module\\

\hline
$L_k $ &  length of  $k^{th}$ link\\

\hline

$\theta_k$  &  angle of $k^{th}$ link with the horizontal axis \\
\hline

$D$  &  Diameter of the pipe (represented as \textbf{\O}) \\

\hline
$\mu$  & coefficient of friction \\

\hline
$f_{x}$  & force acting in $x$ direction \\

\hline
$f_{y}$  & force acting in $y$ direction \\

\hline
$M_J$  &  Moment acting on joint J \\

\hline
$\tau_k$  & torque \\

\hline

\end{tabular}
\label{table:Nomenclature}
\end{table}

\section{Optimization Formulation to find springs' stiffness}
\label{optimization}
\begin{figure}[h!]
\centering
\subfloat[With lower spring stiffness at Joints $J_{1}$ and $J_{4}$ ]{\includegraphics[width=0.12\textwidth,height=0.16\textheight,angle=-0] {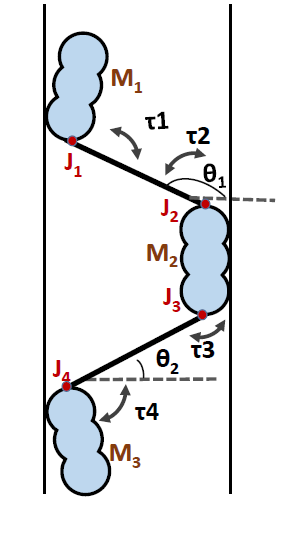}} 
\hspace{0.5cm}
\subfloat[With higher spring stiffness at Joints $J_{1}$ and $J_{4}$ ]{\includegraphics[width=0.12\textwidth,height=0.16\textheight, angle=-0]{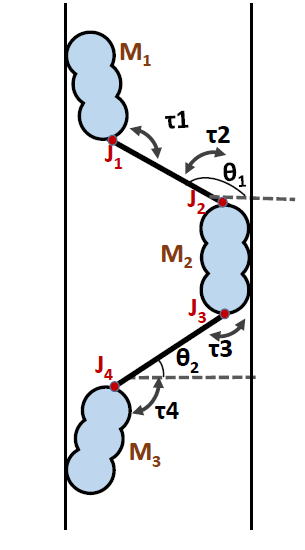}}

\caption{Non-optimal spring stiffness at joints $J_{1}$ and $J_{4}$}
\label{fig:spr_stiff}
\end{figure}

The basic principle behind in-pipe traversal of the robot is generating sufficient traction force by pushing the modules against the surface of the pipe. The preloaded torsion springs at the modules-links joints provide traction force to climb the pipe. While low stiffness torsion springs leads to slippage as a result of insufficient traction force, higher values result in an unnecessarily larger moment at these joints. This is illustrated in Fig. \ref{fig:spr_stiff}. Therefore, an optimal spring stiffness estimation is critical to quasi-statically balance the robot while climbing.

\subsubsection{Objective Function}

The estimation of optimal springs' stiffness is formulated as an optimization problem with an objective to minimize the joint moments, which ensures that the springs are neither too stiff nor too relaxed.   

\begin{align}
\begin{split}
min \sum_{j=1}^{4}|\tau_j|
\end{split}
\end{align}

This function being linear and convex is guaranteed to converge to a global optima. The constraints posed by the geometry, model as well as motion of the robot to perform this optimization, are discussed as follows.

\subsubsection{No-slip constraint}
To avoid slippage at robot-pipe surface interface, friction force which directly relates with the wheel torque must satisfy the following constraints.
\begin{align}
\begin{split}
F_{i}=F_{static} \le \mu N_{i}, \forall i \in \{1,2,3\}
\end{split}
\label{eqn:noslip}
\end{align}
Moreover, the maximum traction force is constrained by the driving motor torque limits.

\begin{align}
\begin{split}
 \le \frac{2\tau_{max}}{(d/2)}
\end{split}
\label{eqn:motortorque}
\end{align}

The unknown parameters $(N_{i}, F_{i}, \tau_{i})$, are determined by formulating a model of the robot which associates all the forces and torques with the spring joint moments. As the robot crawls at a speed of 0.1 m/sec and at such low speed, the motion of a robot is dominated by the surface forces rather than dynamic and inertial effects \cite{c12}, the slow motion crawling behavior can be well captured by the quasi-static model of the robot.

\subsubsection{Quasi-static analysis and Kinematic constraints}
\begin{figure}[t]
\centering
\includegraphics[height=0.2
\textheight,width=0.5\textwidth, angle=0.4]{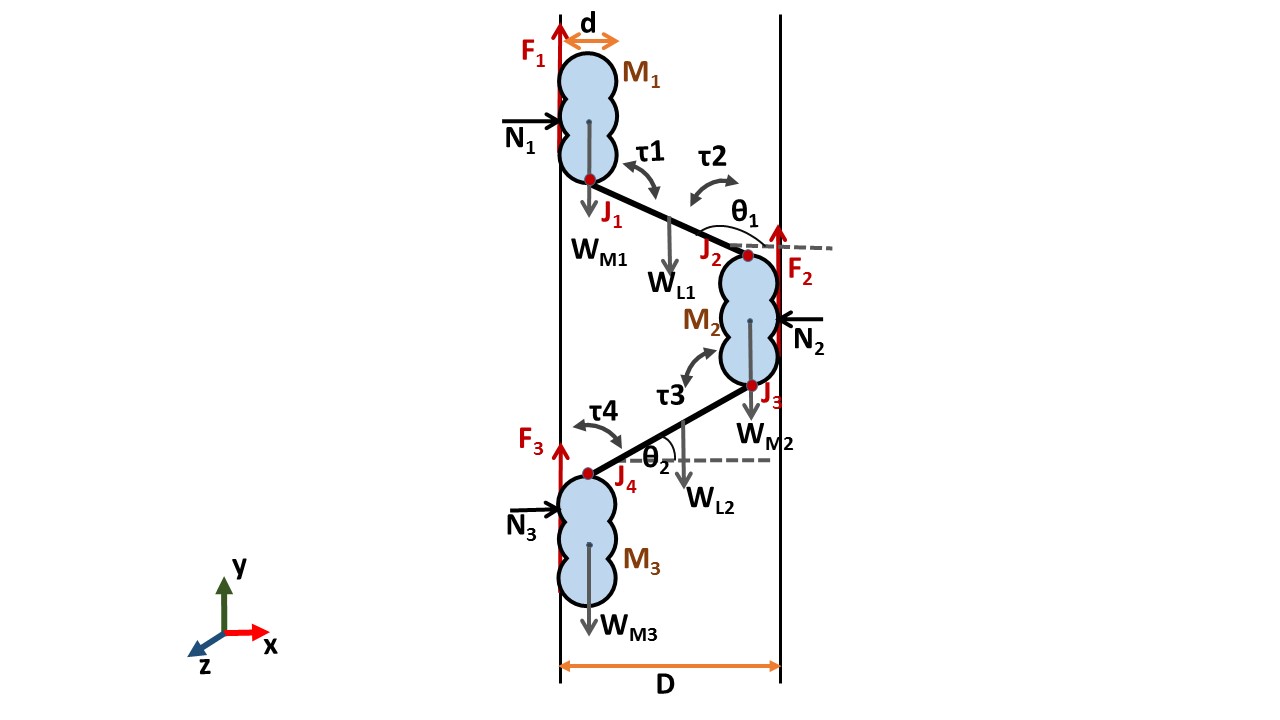}
\caption{Quasi-static configuration in vertical straight pipe}
\label{fig:vert}
\end{figure}

For the mathematical analysis and simulation, each module is modeled as a cascade interconnection of Omniballs, as shown in Fig. \ref{fig:vert}. With the modules aligned in-line with the straight pipes, the joint angles $\theta_1,\theta_2$ are geometrically determined as following.

\begin{align}
\begin{split}
&\theta_1=\pi- cos^{-1} (\frac{D-d}{L_1}) \\ 
&\theta_2=cos^{-1}  (\frac{D-d}{L_2})  
\end{split}
\label{eqn:vert_theta}
\end{align}

The posture parameters ($\theta_1,\theta_2$) are further used to obtain the quasi-static model equations by balancing the forces (equations \ref{eqn:f_x}, \ref{eqn:f_y}) and joint moments at $J_1,J_2,J_3$ and $J_4$ (equations \ref{eqn:MJ1}, \ref{eqn:MJ2}, \ref{eqn:MJ3}, \ref{eqn:MJ4}), as follows.

\begin{align}
\begin{split}
\hspace{-2cm}\Sigma f_x=0, \quad \quad \quad \quad N_1  - N_2 + N_3= 0
\end{split}
\label{eqn:f_x}
\end{align}

\begin{align}
\begin{split}
\Sigma f_y=0, \quad \quad \quad \quad     & F_1 + F_2 + F_3-wm_1-wm_2\\
&-wm_3-wl_1-wl_2=0
\end{split}
\label{eqn:f_y}
\end{align}

\begin{align}
\begin{split}
\hspace{-1cm}\Sigma M_{J_1}=0, \quad \quad \quad F_1d/2+N_1l_1/2-\tau_1=0
\end{split}
\label{eqn:MJ1}
\end{align}

\begin{equation} 
\begin{split} 
\Sigma M_{J_2}=0, \quad  \quad \quad 
&F_1L_1\cos\theta_1+N_1L_1\sin\theta_1-\\
&wm_1L_1\cos\theta1-
wl_1L_1/2\cos\theta_1+\\
&\tau_1-\tau_2=0
\end{split} 
\label{eqn:MJ2}
\end{equation}

\begin{equation} 
\begin{split} 
\Sigma M_{J_3}=0, \quad  \quad \quad 
&-F_2d/2+N_1l_1-N_1l_2/2+\\
&\tau_2-\tau_3=0
\end{split} 
\label{eqn:MJ3}
\end{equation} 

\begin{equation} 
\begin{split} 
\Sigma M_{J_4}=0, \quad  \quad  \quad
&-F_1L_2\cos\theta_2+N_1L_2\sin\theta_2-\\
& F_2L_2\cos\theta_2-N_2L_2\sin\theta_2+\\ 
&(wm_2+wm_1+wl_1)L_2\cos\theta_2+\\ &wl_2(L_2/2)\cos(\theta_2)+ \\
&\tau_3-\tau_4=0; \\
& N_3l_3/2-F_3d/2-\tau_4=0
\end{split} 
\label{eqn:MJ4}
\end{equation} 

These equations form the equality constraints and are represented as $Ax=b$, where
$x$ is a vector of variables ($x=[(F_{i})^T,(N_{i})^T,(\tau_{i})^T]^T$).
 
Therefore, the optimization problem can be formulated as 
\begin{align}
\begin{split}
&\min_{\tau_j} \sum_{j=1}^{4}|\tau_j|\\
\text{subject to} \quad &Ax=b,\\ 
                        &F_{i} \le \mu N_{i}, \quad \forall i \in \{1,2,3\}
\end{split}
\label{eqn:spring_opti}
\end{align}

With the design parameters listed in Table \ref{table:Design Parameters}, the formulated constrained Optimization (eqn.\ref{eqn:spring_opti}) yields a minimal set of passive compliant joint torques at $J_1, J_2, J_3$ and $J_4$ to statically balance the robot, which is further used to obtain the stiffness values, assuming the springs to be linear.

\begin{align}
\begin{split}
 &\tau_i = k(\theta_i-\theta^{initial}_{i}),\\
 \text{where,} \quad &k: \text{spring stiffness}\\
                     &\theta_i: \text{current } i^{th} \text{ joint angle}\\
                     &\theta^{initial}_{i}:\text{initial joint angle (preloaded) of the } i^{th} \text{ joint}
\end{split}
\label{eqn:spring}
\end{align}

\section{Simulation and Experimental results}
\label{results}

All simulations were carried out in ADAMS MSC, a multi-body dynamics simulator to validate the proof of concept of the design with a lumped model of the robot. The springs stiffness values were estimated with the design parameters listed in Table \ref{table:Design Parameters}. 
Also, a prototype of the proposed design was developed and the simulation and numerical results were validated on it.

\subsection{In straight pipes}

In straight pipes (vertical or horizontal), all 3 modules of the robot are aligned in-line with the pipe and the preloaded torsion springs provide the necessary traction force to quasi-statically balance the robot and facilitate slip free driving motion. The robot was manually controlled by an operator. All 3 modules are synchronously driven to propagate the robot in forward/backward direction. 

For vertical straight climbing in \text{\O}75mm acrylic pipes with $\mu=0.7$, the joint angle values obtained from equation(\ref{eqn:vert_theta}) are $\theta_1=115^\circ$ and $\theta_2=65^\circ$ and joint moment values are $\tau_{J_1}=0.2359$ Nm, $\tau_{J_2}=0.3683$ Nm, $\tau_{J_3}=0.2760$ Nm, $\tau_{J_4}=0.1310$ Nm, which result in the following values of springs stiffness at $J_1$,$J_2$,$J_3$ and $J_4$.\\
$k_1$= 0.0096 Nm/deg,\quad
$k_2$ = 0.0056 Nm/deg,\\
$k_3$ = 0.0042 Nm/deg,\quad
$k_4$ = 0.0053 Nm/deg 

\begin{figure}[h!]
\hspace{1cm}
\subfloat[]{\includegraphics[width=0.15\textwidth,height=0.16\textheight]{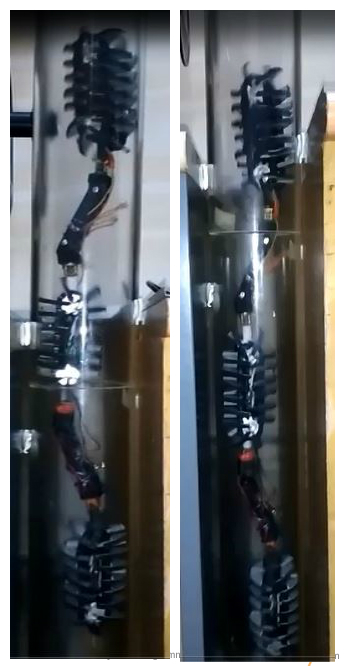}\label{fig:acr}}
\hspace{1cm}
\subfloat[]{\includegraphics[width=0.16\textwidth,height=0.16\textheight]{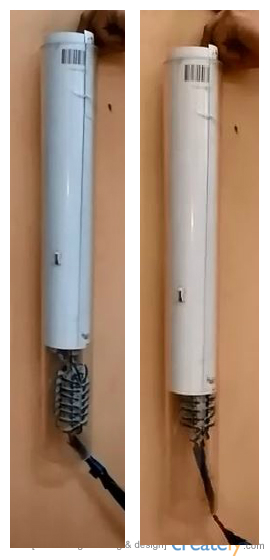}\label{fig:glossy}}
\caption{(a) showing vertical climbing in acrylic pipe($\mu$=0.7); (b) in glossy paper surface($\mu$=0.55)}
\label{fig:friction_vertical}
\end{figure}

With these stiffness values, the robot could successfully climb acrylic vertical pipe as demonstrated in Fig.\ref{fig:acr}. Subsequently, replacing the springs at $J_2$ and $J_3$ with SEAs enables it to climb a pipe with lower friction coefficient ($\mu$=0.55) as depicted in Fig. \ref{fig:glossy}.

Video for climbing in vertical acrylic pipe: \url{https://youtu.be/B7NorlN47MY} \\
Video for climbing in vertical glossy surface: \url{https://youtu.be/BZTHlS3GgTc}

\subsection{In bend pipes}
\begin{figure}[htp]
  \centering
   
   \begin{minipage}[c]{0.2\textwidth}
  \hspace{0cm}  \includegraphics[width=1\textwidth,height=0.14\textheight, angle=-0]{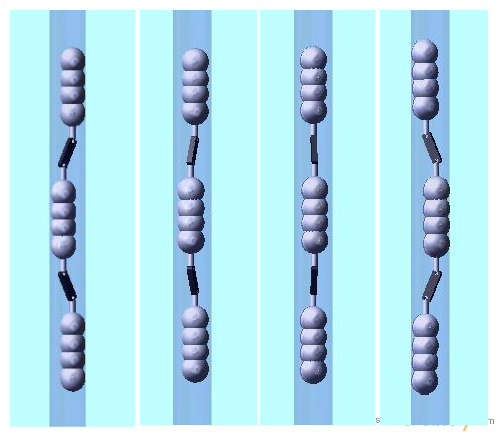}
    \caption{Robot rolls about the axis of the pipe}
    \label{fig:roll}
  \end{minipage}\qquad
  \begin{minipage}[c]{0.2\textwidth}
  \includegraphics[width=1\textwidth,height=0.12\textheight,angle=-0]{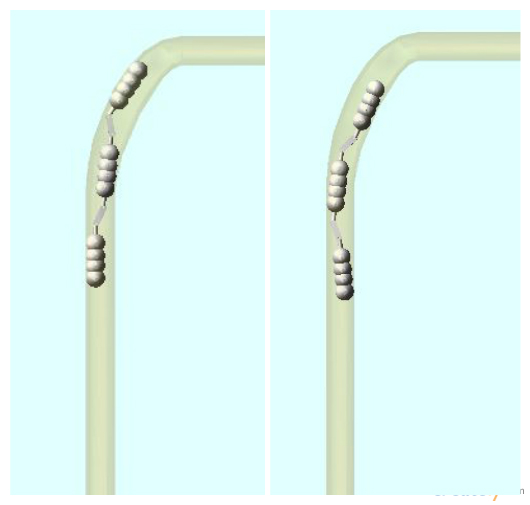}
   \caption{Maximum energy posture (left) and minimal energy posture (right) for turning}
    \label{fig:energy}
  \end{minipage}

\end{figure}

To negotiate bends, the robot must be aligned along the direction of bend before encountering it. The desired alignment is achieved, owing to the sideways rolling motion of OmniCrawler modules, by synchronously rotating all the modules about their own roll axis such that the whole robot body rotates about the axis of the pipe. This is shown in Fig.\ref{fig:roll}. This rolling motion assists it to reach a configuration that overcomes bend in a minimum energy posture where compliant joints apply minimal torque during bending, as shown in Fig.\ref{fig:energy}.

\begin{figure*}
\centering    
\includegraphics[width=1\textwidth,height=0.3\textheight]{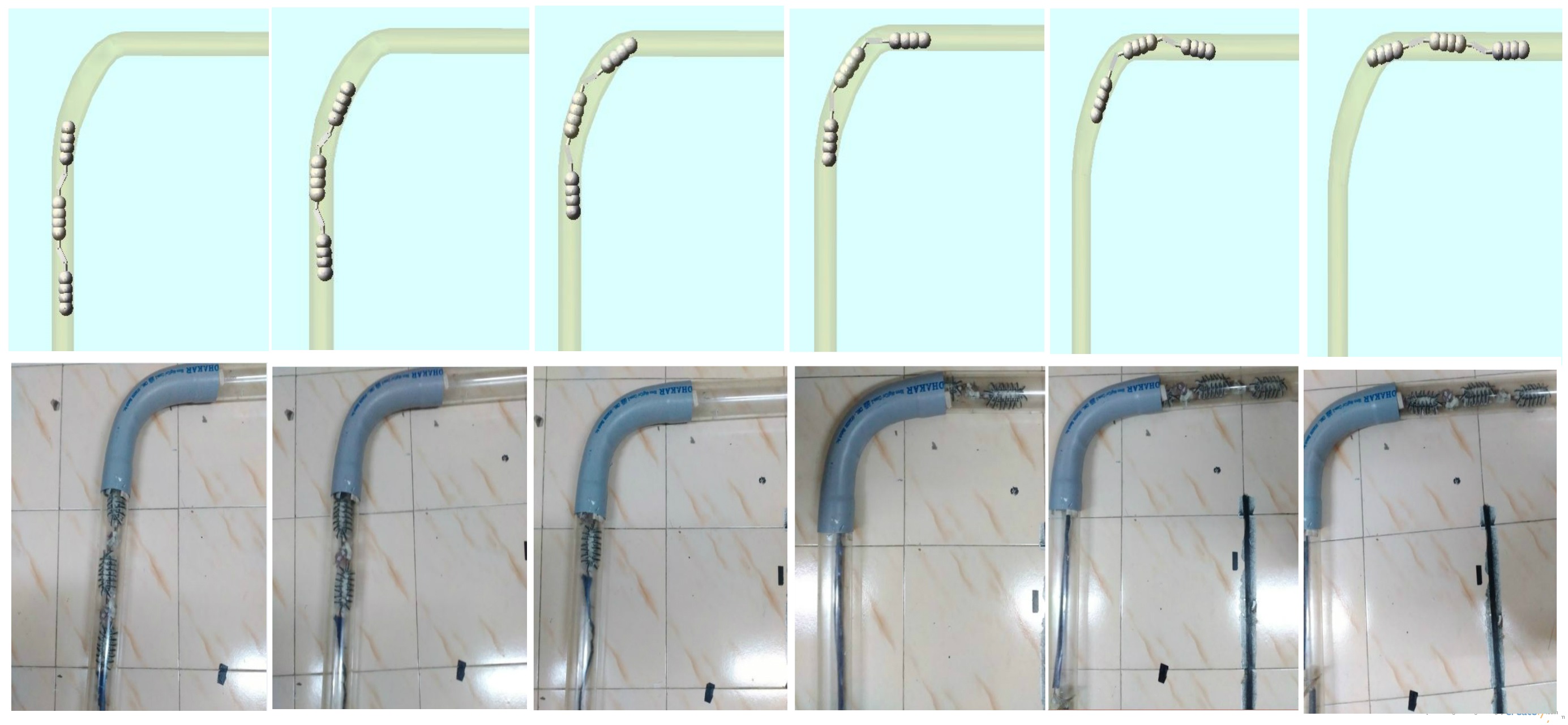}\label{fig:smooth_bend}   
\caption[Optional caption for list of figures 5-8]{Simulation and Experimental results demonstrating robot negotiating smooth 90 degree bend.}    
\label{fig:90_degree}
\end{figure*}

The prototype of the robot with SEAs was tested to negotiate a \text{$\O$}75mm smooth $90^ \circ$ bend in vertical pipe. As the robot traverse through the bend, SEAs are actuated to comply with the pipe turn. Since, the aim of this experiment was to validate the proof of concept of the robot design parameters to overcome smooth $90^ \circ$ bend of a \text{$\O$}75mm pipe, SEAs were not controlled with any estimated joint torque values (at the bends) and the robot was manually controlled by an operator. However, the joint torque estimation at the bend was done in simulation which could later be used for SEA torque control. Fig. \ref{fig:90_degree} demonstrates the simulation and experimental results of the robot steering a $90^\circ$ smooth turn.

\subsection{In T-Junction}
\begin{figure*}
\centering
\hspace{-2cm}
\subfloat[Robot rolls to orient itself such that all 3 modules remain in contact with the surface throughout the motion.]{\includegraphics[width=0.35\textwidth,height=0.15\textheight, angle=-0]{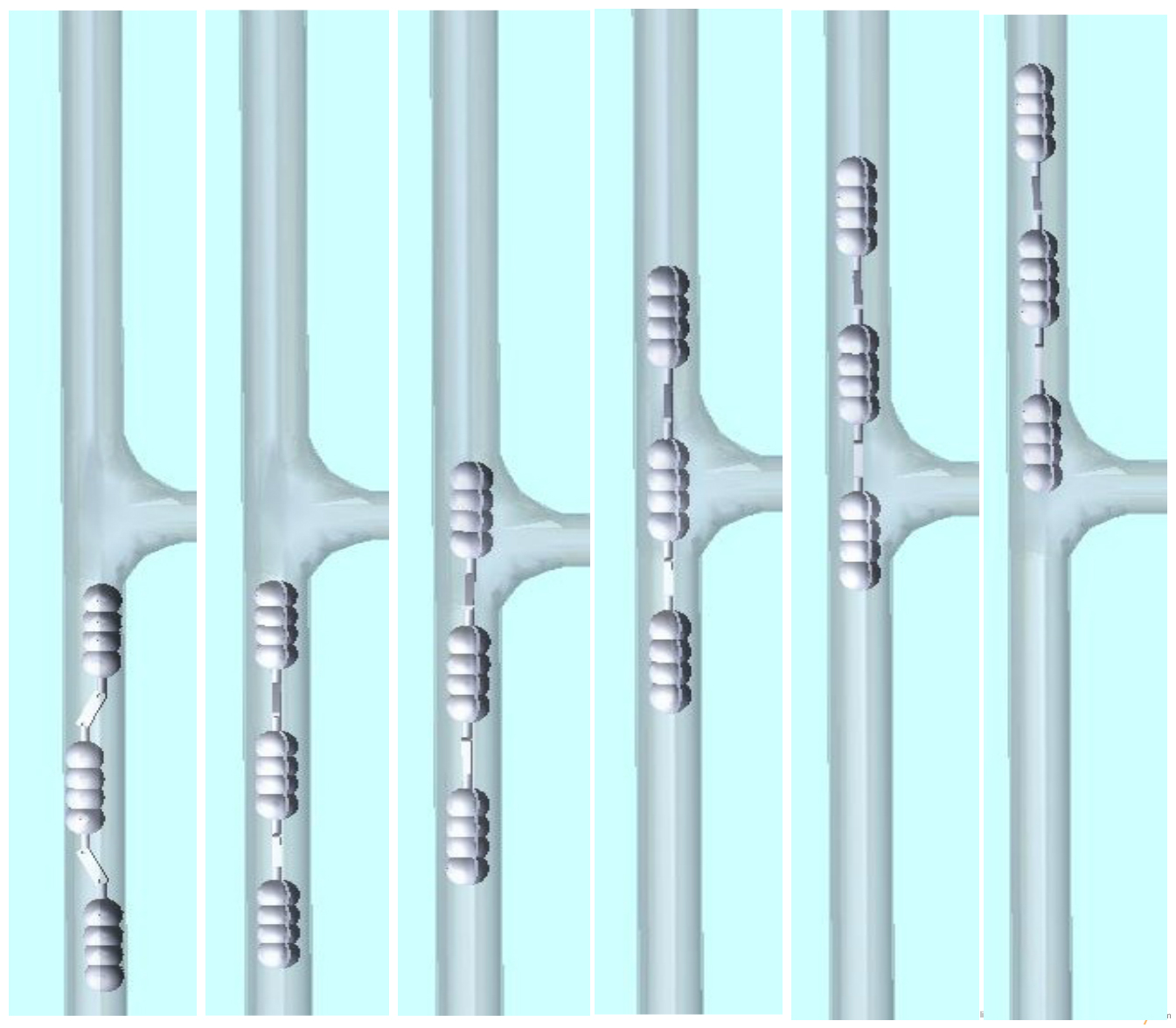}\label{fig:T_1}}
\hspace{2cm}
\subfloat[Robot rolls to orient itself such that 1st module is in contact with surface but the 2nd module loses contact at junction.]{\includegraphics[width=0.35\textwidth,height=0.15\textheight,angle=-0] {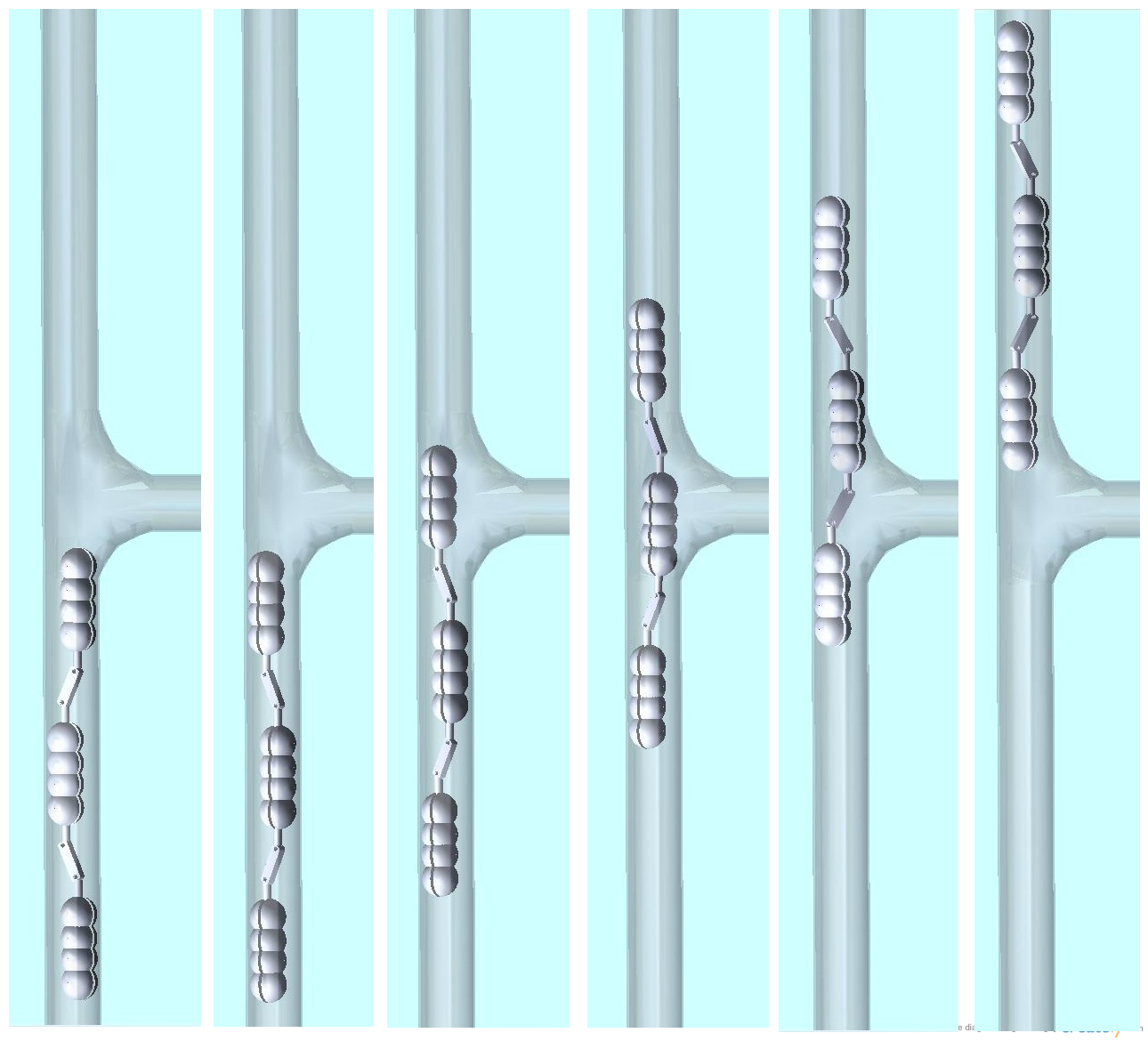}\label{fig:T_2}} 
\caption{Simulation results showing robot overcoming T-junction}
\end{figure*}

Motion singularity often occurs in single module robots while surpassing branches, when it losses contact with the pipe surface and is not able to go through the junction \cite{c11}. The modularity as well as holonomy of the proposed design enable it to avoid motion singularity at junction. The holonomic motion aligns the robot such that none of the 3 modules loses contact with the pipe surface throughout the motion, as shown in Fig. \ref{fig:T_1}. However, in cases where any of the rear modules may lose contact at the junction, the collaborative push/pull forces of other modules overcomes singularity of the module. This is illustrated in Fig. \ref{fig:T_2}, where the robot continues to propagate forward even when 2nd module loses contact with the pipe surface at the junction.

\begin{figure*}[h!]
\centering    
\subfloat[]{\includegraphics[width=1\textwidth,height=0.08\textheight]{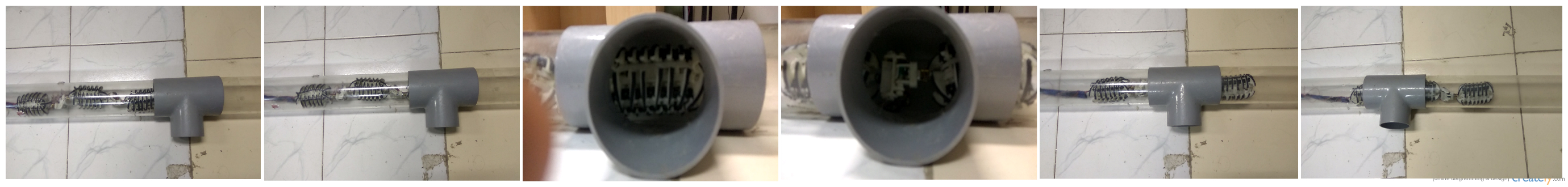}\label{fig:smooth_bend}}    
\caption[Optional caption for list of figures 5-8]{Experimental results demonstrating how the robot overcomes T junction.}    
\label{fig:T_junction}
\end{figure*}

Subsequently, an experiment was carried out in horizontal pipe with T-junction, as shown in Fig. \ref{fig:T_junction}. Since, the size of module is bigger than the size of the junction, robot easily bypass the junction without getting stuck there. 

\section{Conclusion and Future Work}
In this paper, a novel compliant modular Omnicrawler based in-pipe robot has been proposed. The holonomic motion of each module enables the alignment of robot to overcome bends and obstacles. For a given pipe environment, a set of optimal springs stiffness values was calculated to quasi-statically balance the robot in vertical pipes and was experimentally validated. Furthermore, the SEAs were incorporated at 2 of the 4 joints, to overcome smooth bends and friction coefficient variations in pipes. The proposed design has the capability to climb in small diameter pipes. However, the sharp pipe turns was not addressed by this design. Therefore, our future work would focus on modification of the design in order to negotiate sharp turns. Additionally, a closed loop torque control strategy for the SEA could be implemented to cover a wide range of pipe diameter and friction coefficient variations.

\end{document}